\documentclass[a4paper, 10pt, conference]{ieeeconf} 
\IEEEoverridecommandlockouts  

\makeatletter
\let\NAT@parse\undefined
\makeatother

\usepackage{graphicx}

\usepackage{chngpage}	
\usepackage{tabularx}	
\usepackage{amsmath}
\usepackage{amssymb} 
\usepackage{color}
\usepackage{float}
\usepackage{calc}
\usepackage{multicol}
\usepackage[small]{caption}
\usepackage[bookmarks,bookmarksnumbered,%
    citebordercolor={0.8 0.8 0.8},filebordercolor={0.8 0.8 0.8},%
    linkbordercolor={0.8 0.8 0.8},pagebordercolor={0.8 0.8 0.8},%
    urlbordercolor={0.8 0.8 0.8},%
    ]{hyperref}

\usepackage{cite}
\usepackage{natbib}
\usepackage{booktabs}
\usepackage{algorithm, algorithmic}
\usepackage{subfigure}
\usepackage{wrapfig}
\graphicspath{{./figures/}}
\usepackage{rotating}
\usepackage{multirow}

\usepackage{lineno}
\usepackage{colortbl}

\usepackage{comment}

\newcommand{\vect}[1]{\boldsymbol{#1}}
\DeclareMathOperator*{\argmax}{argmax}

\begin{document}

\title{Self-Learning for Player Localization in Sports Video
}


\author{Kenji Okuma,  
         David G. Lowe, 
         James J. Little\\
         Department of Computer Science, The University of British Columia.\\
\tt{\{okumak, lowe, little\}@cs.ubc.ca}}

\maketitle

%
%
\begin{abstract}
This paper introduces a novel self-learning framework that automates the label acquisition process for improving models for detecting players in broadcast footage of sports games.  Unlike most previous self-learning approaches for improving appearance-based object detectors from videos, we allow an unknown, unconstrained number of target objects in a more generalized video sequence with non-static camera views.  Our self-learning approach uses a latent SVM learning algorithm and deformable part models to represent the shape and colour information of players, constraining their motions, and learns the colour of the playing field by a gentle Adaboost algorithm.  We combine those image cues and discover additional labels automatically from unlabelled data.  In our experiments, our approach exploits both labelled and unlabelled data in sparsely labelled videos of sports games, providing a mean performance improvement of over 20\% in the average precision for detecting sports players and improved tracking, when videos contain very few labelled images.
\end{abstract}

\section{Introduction}
Recent advances in object detection have enabled computers to detect many classes of objects, such as faces, pedestrians, and cars.  Modern digital cameras and video conferencing systems often have a built-in face detection system to automatically focus on faces.  Pedestrian detection has been employed for monitoring surveillance videos and supporting safer driving of cars.  However, these machine learning methods suffer from a major drawback --- they require a large amount of training data. In order to achieve performance levels that are high enough for practical commercial applications, it is common that more than a million labelled instances are used for the training, which must be acquired at great expense.

One way to resolve this issue is to employ abundant unlabelled data.  Active learning has been adopted to train object detectors without much human effort \citep{Okuma2011, Vijayanarasimhan2011}. With abundant unlabelled data, crowdsourcing is also a powerful tool to utilize human labour efficiently with reduced cost for obtaining abundant labels.  LabelMe \citep{Russell_IJCV2008} and other interactive user interfaces on Amazon Mechanical Turk such as one by \citep{Sorokin2008} and the Visipedia project \citep{Welinder2010} address inexpensive acquisition of labels from a large pool of thousands of unlabelled images.  Recently, crowdsourcing has also been utilized for annotating a collection of video data.  Interactive annotation tools on the Web such as VATIC, a video annotation tool by \citep{Vondrick2010}, and LabelMe video \citep{Yuen2009} have become publicly available in the computer vision community to foster large scale labelling of unlabelled video data.  However, those crowdsourcing tools are designed primarily for reducing the overall labelling cost in terms of time and money.  They consider neither the impact of each label for improved performance of a classification model nor reducing the size of training data.  

Another way to resolve the shortage of labelled data is to exploit both labelled and unlabelled data.  There has been, especially in recent years, a significant interest in semi-supervised learning, which exploits both labelled and unlabelled data to efficiently train a classifier.  Semi-supervised learning approaches have shown success in various domains such as text classification \citep{Nigam_ML2000}, handwritten digits recognition \\ \citep{Lawrence2005}, track classfication \citep{Teichman_IJRR2012}, and object detection \citep{Rosenberg2005, Leistner2007, Ali2011, Siva2012, Yao2012a}.  There is a large literature on methods of semi-supervised learning, which originally dates back to the work of Scudder \citep{Scudder1965}.  

In this paper, we use semi-supervised learning for improving an appearance-based model of target objects.  Most of the recent approaches \citep{Leistner2007, Ali2011, Yao2012a} exploit a relatively small amount of labelled data to discover a meaningful portion of training samples for improving object localization in video sequences.  None of these approaches, however, address the use of video data with non-stationary camera views. Combined motions from both a non-stationary camera and moving target objects cause inherent localization difficulties.  We show that our approach improves player localization on broadcast footage of sports, which allows an unknown, unconstrained number of target objects in more generalized video sequences with non-static camera views.  For improving player localization, we address how to maximize the impact of labels by selecting examples that are most likely to be misclassified by the current classification function, and to reduce the overall labelling cost by making the labelling process fully automatic.  

\section{Weakly-supervised self-learning for player localization}
Given sparsely labelled video data that consists of $n$ different video sequences $\{V_i\}_{i=1}^{n}$ where each sequence contains a different number of image frames, $V_i = \\ \{\vect{x}_1, \dots, \vect{x}_{n_i}\}$, the task is to train an initial model $H: \mathcal{X} \mapsto \mathcal{Y}$ from a small set of labels $\mathcal{L} = \{(\vect{x}_1, y_1), \dots, \\ (\vect{x}_l, y_l)\}$ and exploit additional unlabelled data $\mathcal{U} = \\ \{\vect{x}_{l+1}, \dots, \vect{x}_{l+m}\}$ for improving the model, assuming that $\vect{x} \in \mathcal{X}$, $y \in \mathcal{Y}$, and $l \ll m$.  In this paper, we will use hockey and basketball video data for learning an appearance-based model of sports players. This can be viewed as a weakly-supervised learning problem because we deal with videos without localization of the target objects.  Unlike most previous semi-supervised learning methods, we allow an unconstrained, unknown number of players that appear in each frame of a video sequence.    

We propose to use self-learning, which is one of the most commonly used semi-supervised learning methods \citep{Chapelle2006}, to lower the requirement for extensive labelling.  Self-learning is a wrapper algorithm that repeatedly uses a supervised learning method.  It starts with a small set of labels to train the initial model.  In each iteration, the model is used to evaluate unlabelled data and to obtain predictions.  The model is then retrained with a selected portion of predictions as additional labels. This process is repeated until some stopping criterion is met. 
\begin{figure*}[ht!]
	\center
	\includegraphics[width=0.8\textwidth]{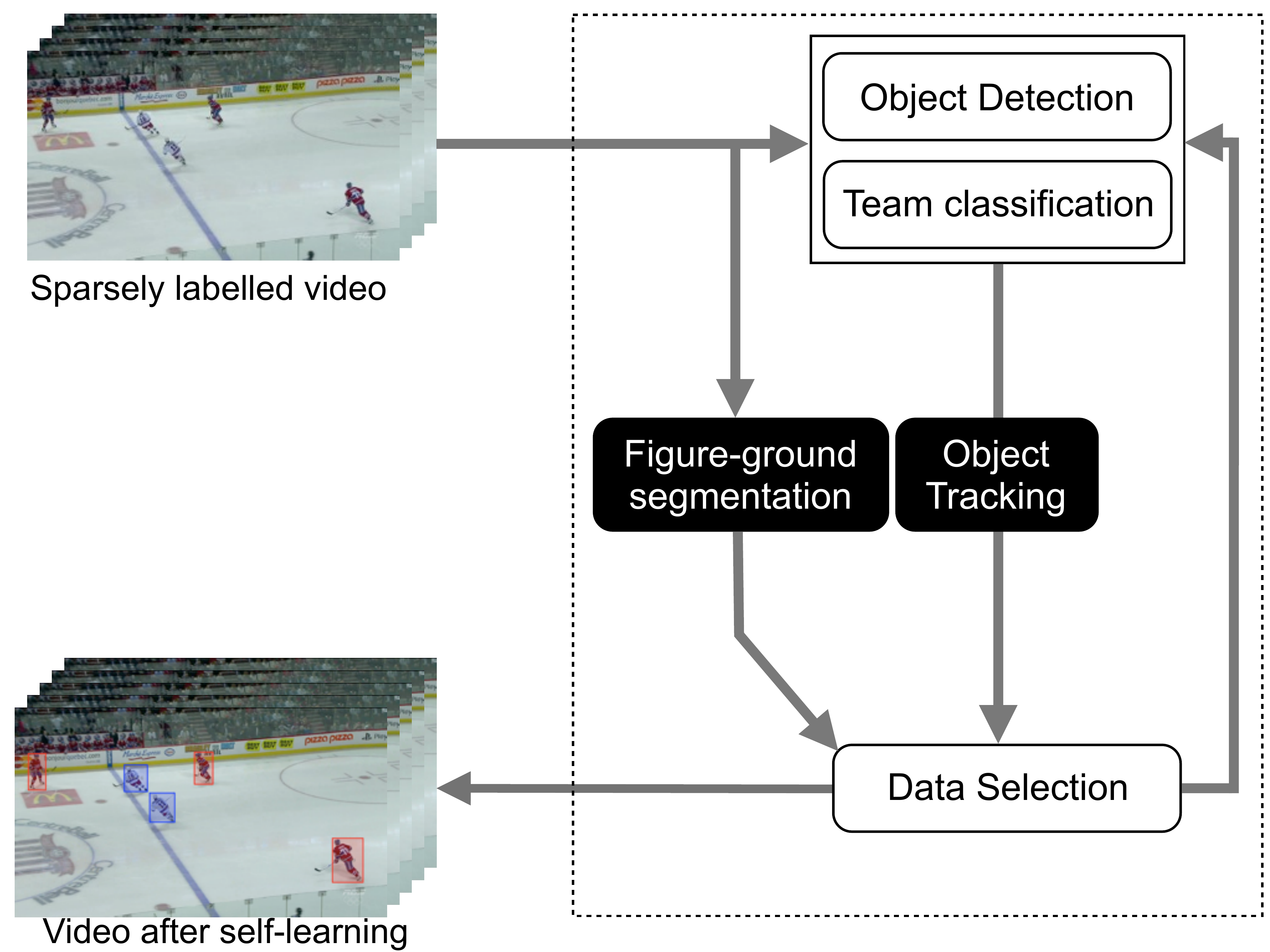} 
  	\caption[System overview of our self-learning framework]
	{{\bf System overview of our self-learning framework.}
	{\rm\quad Black boxes mean that models are not updated during the training process and treated as a black box.  The system takes a sparsely labelled video with a small set of fully labelled image frames as input and trains initial classification models.  Our self-learning approach uses these models to explore the unlabelled portion of data, collecting additional training labels, and updates these models for improved performance.  This process is repeated multiple times and produces a more complete set of labels in colour-specific tracklets in the video.
	}}
\label{fig:system_overview}
\end{figure*}

Our self-learning system has several stages as shown in \autoref{fig:system_overview}.  The training procedure starts from initializing with a small set of labelled images and a large set of unlabelled images from sparsely labelled video data.  Then the system iterates over the following steps.  First, a small set of labelled data is used to train initial part-based models for detecting players and classifying their team colour (\autoref{sec:player_detection} and \autoref{sec:colour_classification}).  Second, these appearance-based models are applied to the unlabelled data and generate a set of detection bounding windows.  Third, these bounding windows are linked by a Kalman filter and generate a set of tracklets (\autoref{sec:player_tracking}).  Finally figure-ground segmentation is applied to validate these tracklets.  The resulting set of validated tracklets is used as additional labels to re-train current classification models.  Algorithm \autoref{alg:self_learning} summarizes this process. 
\begin{algorithm}[ht!]	
	\caption{{\bf: Self-learning for player localization in videos} \\ 
	 Given training video sequences $\{V_i\}_{i=1}^{n}$, randomly select $m$ labelled images that contain an initial set of labelled data $\mathcal{L} = \{(\vect{x}_1, y_1, c_1), \dots , (\vect{x}_l, y_l, c_l)\}$ where $\vect{x}$ is a window descriptor, $y$ is a class label, and $c$ is a team colour label. The number of self-learning sessions is set as $n_s=5$.}	
	\begin{algorithmic}[1]		
		\STATE {{\bf Initialize} $\mathcal{U}$ with all image frames that are unlabelled in $\{V_i\}_{i=1}^{n}$.}
		\FOR {$n_s$ self-learning sessions}	
		\STATE {{\bf Training classifiers}: \\ Given labelled data $\mathcal{L}$, train a player detector (\autoref{sec:player_detection}) and colour classifiers (\autoref{sec:colour_classification})}		
		\STATE {{\bf Player detection and team classification}: \\ Run the trained classifiers for unlabelled data $\mathcal{U}$ (\autoref{fig:hockey_det_col_results})}
		\STATE {{\bf Player tracking}: \\ Run a Kalman filter to link detection bounding windows (\autoref{sec:player_tracking})}			
		\STATE {{\bf Data selection}: \\ Select a new dataset $\mathcal{L}_{new}$ (\autoref{sec:data_selection}) and add to existing data: $\mathcal{L} = \mathcal{L} \cap \mathcal{L}_{new}$}		
		\ENDFOR
	\end{algorithmic}	
	\label{alg:self_learning}		
\end{algorithm}

\begin{figure*}[ht!]
\center
\begin{tabular}{ c @{\hspace{1mm}} c  c @{\hspace{1mm}} c @{\hspace{1mm}} c  c @{\hspace{1mm}} c  c}
	\includegraphics[width=0.1\textwidth]{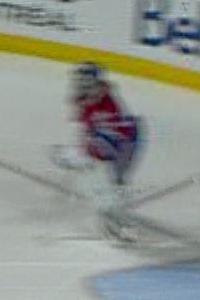}  &
	\includegraphics[width=0.1\textwidth]{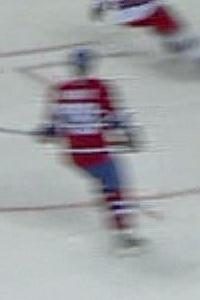}  &
	\includegraphics[width=0.1\textwidth]{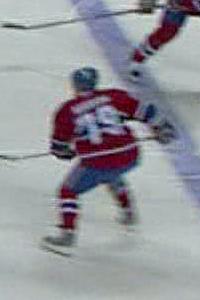}  &
	\includegraphics[width=0.1\textwidth]{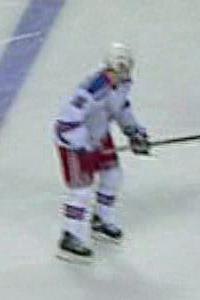}  &
	\includegraphics[width=0.1\textwidth]{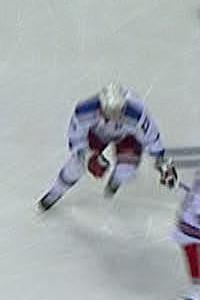}  &
	\includegraphics[width=0.1\textwidth]{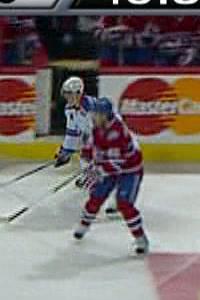}  &
	\includegraphics[width=0.1\textwidth]{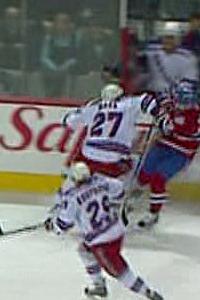}  &
	\includegraphics[width=0.1\textwidth]{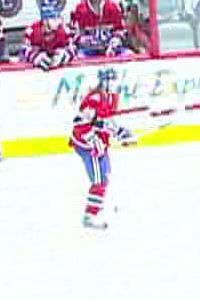}  \\		
	\multicolumn{2}{c}{motion blur} &
	\multicolumn{3}{c}{object pose} &
	\multicolumn{2}{c}{occlusion} &
	illumination \\
\end{tabular}
  \caption[Challenges in player detection]
	{{\bf Challenges in player detection.}  These include motion blur, wide pose variation, occlusion, and sudden illumination change.
	{\rm\quad  }}
\label{fig:hockey_players}
\end{figure*}

There are several reasons why we particularly focus on sports player detection in sports videos. Sports videos are highly structured because the domain knowledge is quite specific (e.g, team colours, the player uniform, the colour of the playing field). But they are still challenging enough to be an interesting problem.  For example, \autoref{fig:hockey_players} shows several major challenges for detecting hockey players. Videos in sports --- especially team sports such as hockey (6 on-field players per team), basketball (5 on-field players per team), and soccer (11 on-field players per team) --- are a rich source of labels for learning the appearance of sports players since each frame of a video almost always contains multiple labels.  Furthermore, accurate localization of sports players is a fundamental requirement for tackling other interesting problems such as action recognition and player recognition.  To the best of our knowledge, our work is the first large scale study of a self-learning framework for learning the appearance of sports players in broadcast footage. 
\section{Semi-supervised learning in videos}
Many algorithms in semi-supervised learning assume that the unlabelled data are independent samples.  However, in a video sequence, the trajectory of object instances, defined by the location of the bounding windows, suggests the spatio-temporal structure of subsequent labels. 

In order to exploit the dependent structure of the video data, several tracking-by-detection approaches \\ \citep{Kalal2010, Leistner2011, Babenko2009} have been proposed to learn an appearance model of an object from videos.  These approaches have the stringent assumption of having only one instance of the target object class in each frame of a video sequence.  Such an assumption strictly limits applications to detection of a single instance of the target object class, where an instance with the highest confidence is identified as a positive label and all remaining instances are labelled as negative. For learning the appearance of an object class such as pedestrians or faces, videos that contain multiple pedestrians in each frame are much more effective than videos with one person in each frame, because they capture occlusion relationships that are not present in single object videos.  But localization of multiple target objects remains difficult, and it prevents most tracking-by-detection approaches from exploiting unlabelled data that are available from such videos. Nonetheless, there are a few approaches that have considered exploiting unlabelled video data with multiple target objects such as \citep{Ramanan2007, Ali2011}.

\citet{Ramanan2007} proposed a semi-supervised method for building a large collection of labelled faces from archival video of the television show Friends.  Their final collection contains 611,770 faces.  Their approach used the Viola {\it et al.}'s face detector to detect faces, grouping them with colour histograms of body appearance (i.e, hair, face, and torso) and tracking them using a part-based colour tracker for multiple people in videos.  Although their approach is effective with large scale data, they performed only one iteration of exploring the unlabelled data for building a large collection of faces and never used the acquired collection for improving the classifiers they used.   

Recently, \citet{Ali2011} implemented self-learning on sparsely labelled videos, which allows any number of instances of the target object class.  
The approach described in \citep{Ali2011} is most related to our approach.  But it uses a different learning approach and has a number of limitations that we address. It has the major limitation that an appearance of target objects must have a single scale where we need to improve player localization for sports players with various sizes.  Furthermore, it assumes a simpler form of video input that could not be applied to broadcast footage of sports.  Their model is based on a rather simple, smooth motion of walking pedestrians in their surveillance data of a stationary camera view.  Sports players have much more complicated, unpredictable motions with more frequent, complex interactions.  Secondly, their approach differs significantly from ours.  They used simple edge based features for representing the shape of pedestrians and used a boosting algorithm and linear programming to exploit the temporal coherence of videos.   We adopt a latent SVM formulation for learning the shape and colour of sports players who have a variety of different poses (i.e., running, jumping, walking, and etc).  We use Kalman filters to link a sparse set of detection boxes, and use figure-ground segmentation as additional information to validate the unlabelled data.  Our work is the first to apply self-learning to videos which contain multiple target objects of a moving camera view.   

\section{Player detection}
\label{sec:player_detection}
In order to detect hockey players, we adopt the recent latent SVM (LSVM) approach of \citet{Felzenszwalb_PAMI2009}.  The goal of a supervised learning algorithm is to take $n$ training samples and design a classifier that is capable of distinguishing $M$ different classes.  For a given training set $(\vect{x}_1, y_1), \ldots, (\vect{x}_n, y_n)$ with $\vect{x}_i \in \Re^{N}$ and $y_{i} \in \{-1, +1\}$ in their simplest form with two classes, LSVM is a classifier that scores a sample $\vect{x}$ with the following function,
\begin{equation}
	f_{\vect{\beta}}(\vect{x}) = \max_{\vect{z} \in \vect{Z}(\vect{x})}\vect{\beta} \cdot \Phi(\vect{x}, \vect{z})
\end{equation}
Here $\vect{\beta}$ is a vector of model parameters and $\vect{z}$ are latent values.  The set $\vect{Z}(\vect{x})$ defines possible latent values for a sample $\vect{x}$. Training $\vect{\beta}$ then becomes the optimization problem. 
We approximate the posterior probability $P(y=1|\vect{x})$ of the decision function in a parametric form of a sigmoid \citep{Platt2000, Lin2003}.
\begin{equation}
    \begin{aligned}
	P(y=1|\vect{x}) \approx P(y=1|f) & = \frac{1}{1 + \exp(fA + B)} \\ 
	& \hspace{2mm}{\rm where}\hspace{2mm} f = f_{\vect{\beta}}(\vect{x}) \\
    \end{aligned}
\end{equation}
We used their code for detection and augment it with a colour classifier as described below.
\section{Team classification}
\label{sec:colour_classification}
Our shape-based deformable part model (DPM) gives a tight bounding window of the object (i.e., a hockey player) as well as a set of smaller bounding windows of its corresponding parts.  Given these bounding windows as prior knowledge, the model learns a colour classification function based on deformable parts with the following function:
\begin{equation}
	f_{\vect{\gamma}}(\vect{x}) = \vect{\gamma} \cdot \Phi(\vect{x}, \vect{z_\beta})
\end{equation}
where $\vect{\gamma}$ is a vector of model parameters and $\vect{z_\beta}$ are latent values specified by the shape-based DPM detector.  Following \citep{P'erez2002, Okuma2004, Lu_IMAVIS2009}, we use Hue-Saturation-Value (HSV) colour histograms.  Thus, a feature vector $\vect{x}$ is composed of a set of HSV colour histograms, each of which has $N = N_hN_s+N_v$ bins and corresponds to a unique part of the deformable part models.  A distribution $K(\vect{R}) \triangleq \{k(n;\vect{R})\}_{n=1,\ldots,N}$ of the colour histogram in a bounding window $\vect{R}$ is given as follows:
\begin{equation}
	k(n;\vect{R}) = \eta \sum_{d \in \vect{R}}\delta[b(\vect{d}) - n]
\end{equation}
where $d$ is any pixel position within $\vect{R}$, and $b(\vect{d}) \in \{1,\ldots,N\}$ as the bin index.  $\delta$ is the delta function.  We set the size of bins $N_h$, $N_s$, and $N_v$ as 10.  The normalizing constant $\eta$ ensures that all the bin values are $[0, 1.0]$.  It is important to note that $K(\vect{R})$ is not a probability distribution and is only locally contrast normalized\footnote{In our experiments which are not shown here, we tested our classification model with the distribution of the colour histograms which are normalized to be probability distributions.  However, results were much worse than ones with local contrast normalization\label{fn:repeat}.}, $\max{K(\vect{R})} = 1.0$. 
\begin{figure*}[htbp]	
	\center
	\begin{tabular}{ cc }
		\includegraphics[width=0.48\textwidth]{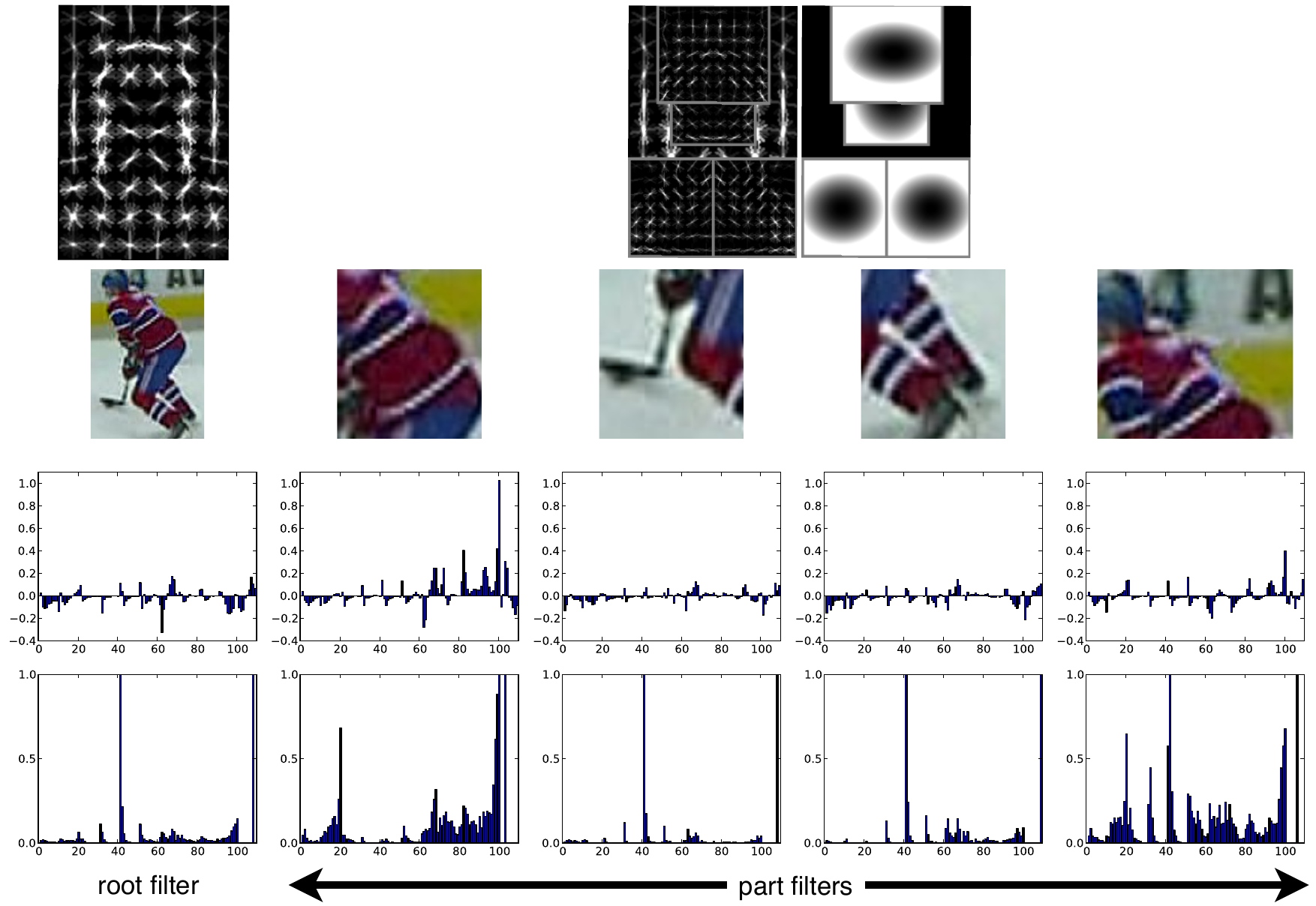}  &
		\includegraphics[width=0.48\textwidth]{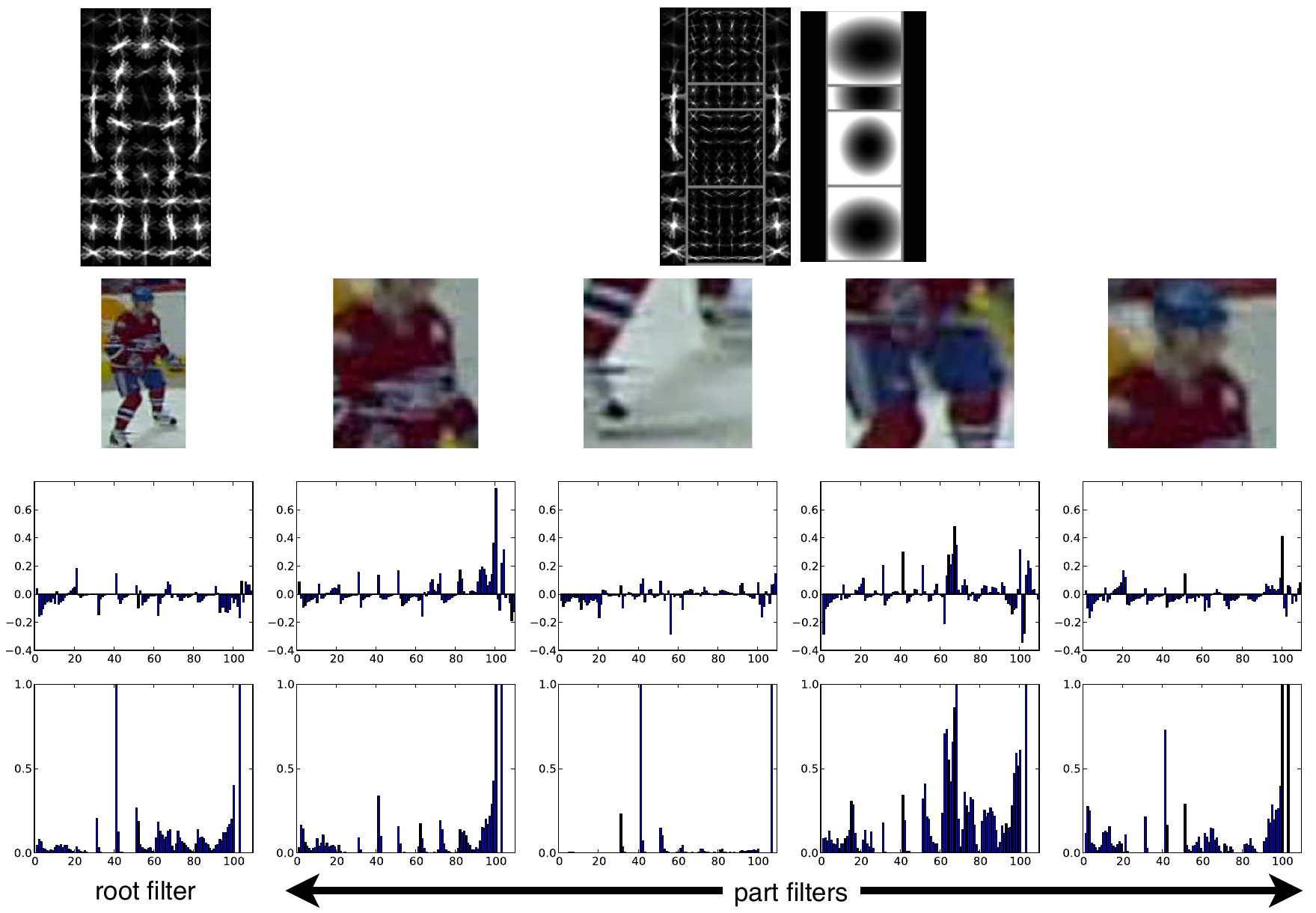} 
	\end{tabular}
	  \caption[Colour-based deformable part models]
		{{\bf Colour-based deformable part models.}
		{\rm\quad  This shows a mixture of two part-based colour models for the Montreal Canadiens team.  For each model, the top row shows the root filter, part filters, and deformation model.  The second row shows corresponding image regions of the object.  The distribution of their learned weights and HSV colour histograms are shown respectively in the third and forth row.  Note noticeably higher weights on those parts that are particularly \emph{discriminative} for classification (e.g., the 2nd column in the left, the 2nd and 4th in the right)}}
	\label{fig:cdpms}	
\end{figure*}
We train a colour model for each team label: ``MTL'' for Montreal Canadiens, ``NYR'' for New York Rangers, and ``ref'' for referees.  \autoref{fig:cdpms} shows two component deformable part models for the Montreal Canadiens team.  The posterior probability of the decision function for each colour classification model is approximated by fitting a sigmoid function \citep{Platt2000, Lin2003}.  Finally, our team colour classification function is formulated as the maximum likelihood of three binary colour classification models.  
\begin{equation}
    y^{*} = \argmax_{y \in \mathcal{Y}} P(y|\vect{x}, \vect{z_\beta})
\end{equation}
where $y$ is a team label and $\mathcal{Y} = \{\text{``MTL''}, \text{``NYR''}, \text{``ref''}, \\ \text{``others''}\}$.  These part-based colour models are highly discriminative since they use the learned latent values $\vect{z_\beta}$ (i.e., location and size of multiple parts of an object) based on the shape-based DPM detector.  Furthermore, these colour models are efficiently trained without optimizing over a large space of latent values, which is the bottleneck of training the latent SVM.    

\begin{figure*}[ht!]
\center
\begin{tabular}{ c c }
	\includegraphics[width=0.45\textwidth]{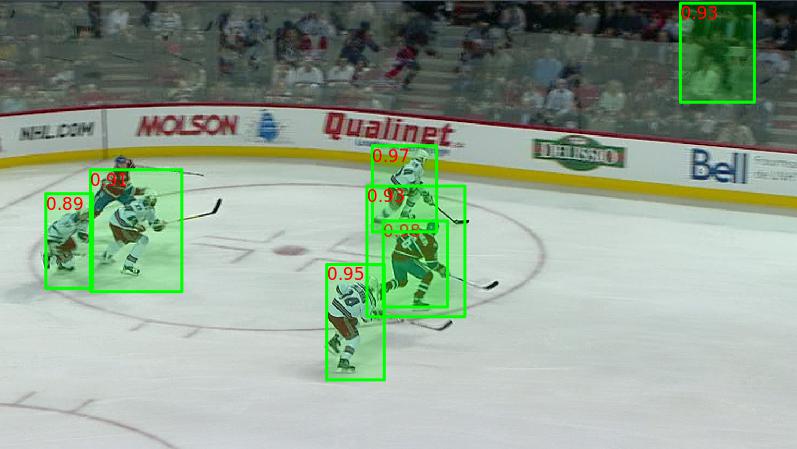}  &
	\includegraphics[width=0.45\textwidth]{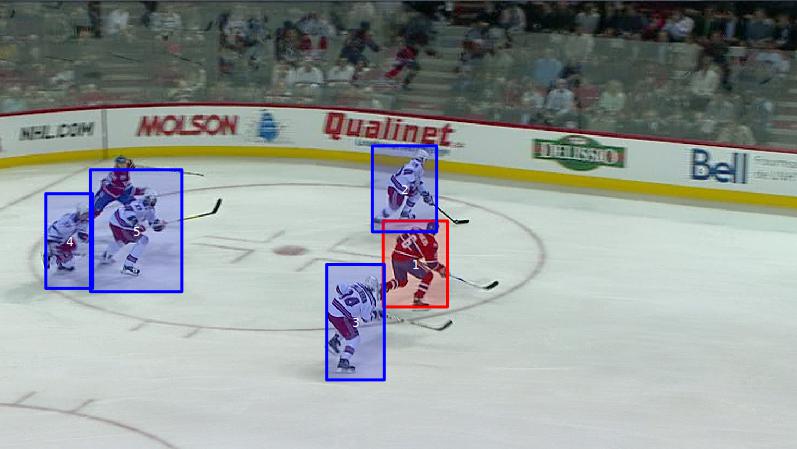}  \\
	(a) detection & (b) detection + team colour classification \\
\end{tabular}
  \caption[Player detection and team colour classification results]
	{{\bf Player detection and team colour classification results.}
	{\rm\quad  This is best viewed in colour.  This shows results of player detection and team colour classification.  Detection bounding windows are shown in green boxes in (a) with their detection confidence in the upper left corner of these bounding windows, and with team colour classification in red and blue boxes in (b). Note that team classification suppresses false positive detections in the background.}}
\label{fig:hockey_det_col_results}
\end{figure*}

For team colour classification, part-based colour models are particularly effective when two teams, the Montreal Canadiens and the New York Rangers, have a similar distribution of colours (e.g., red and blue) in their uniform  (\autoref{fig:hockey_players}).  \autoref{fig:cdpms} shows how multi-part weighted histograms preserve the spatial information of colour distributions, where a single holistic representation cannot. In the figure, there are two different part-based colour models for the Montreal Canadiens, where each model has weighted multi-part colour histograms.  Parts with more discriminative colour are learned to have higher weights.  \autoref{fig:hockey_det_col_results} shows results of team colour classification, which improves detection results of the shape-based model by suppressing those detection windows that do not have the learned team colour labels.  In this case, we had 79\% precision and 57\% recall without team classification (a) and 89\% precision and 54\% recall with team classification by suppressing false positive detection windows (b). 

\section{Figure-ground segmentation and player tracking}
\label{sec:player_tracking}
We developed an interactive labelling tool to learn a figure-ground segmentation model based on a boosting algorithm.  Given a small set of manually labelled foreground pixels and background pixels on the first image, we used the OpenCV implementation of Gentle Adaboost to learn a set of 150 weighted decision trees\footnote{Learning and inference of the model can be further sped up by using decision stumps (i.e., one level decision trees) instead of multi-level decision trees, or reducing the number of weak features.} where the maximum depth of these trees is 10.  We then use the initial model on an additional few images, interactively labelling wrongly classified pixels and update the model with these additional labels.  The process is repeated a few times with no more than 5 images.  

We also tested a saliency measure called ``objectness'' \citep{Alexe2010} because it has been used in state-of-the-art weakly supervised approaches for localizing generic objects.  However, ``objectness'' did not work well in a hockey video mainly due to a small size of hockey players and weak contrast of the colour of hockey players and the rink.  
\begin{figure*}[ht!]
\center
\begin{tabular}{ c c c c}
	\includegraphics[width=0.22\textwidth]{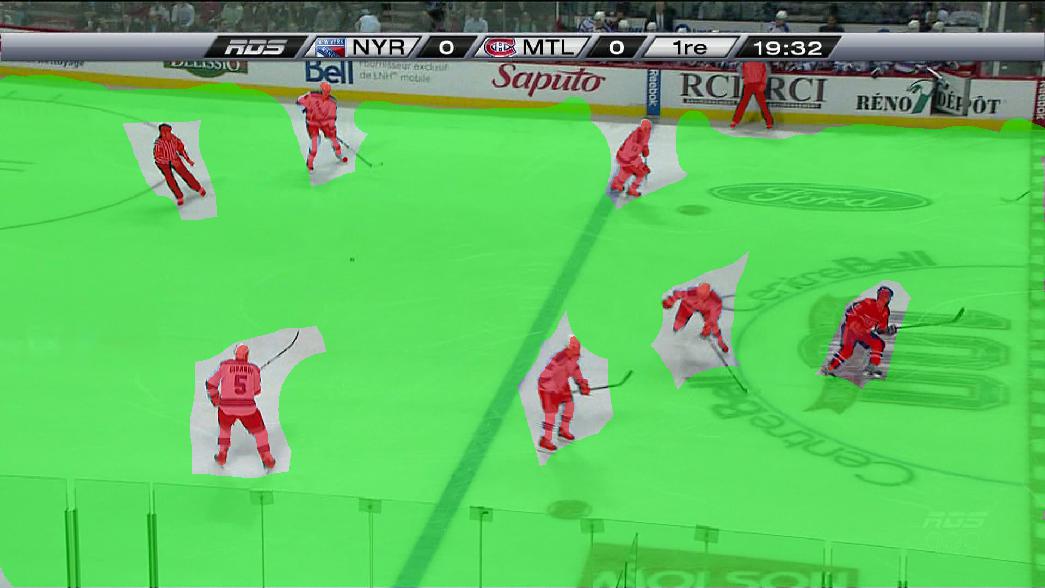}  &
	\includegraphics[width=0.22\textwidth]{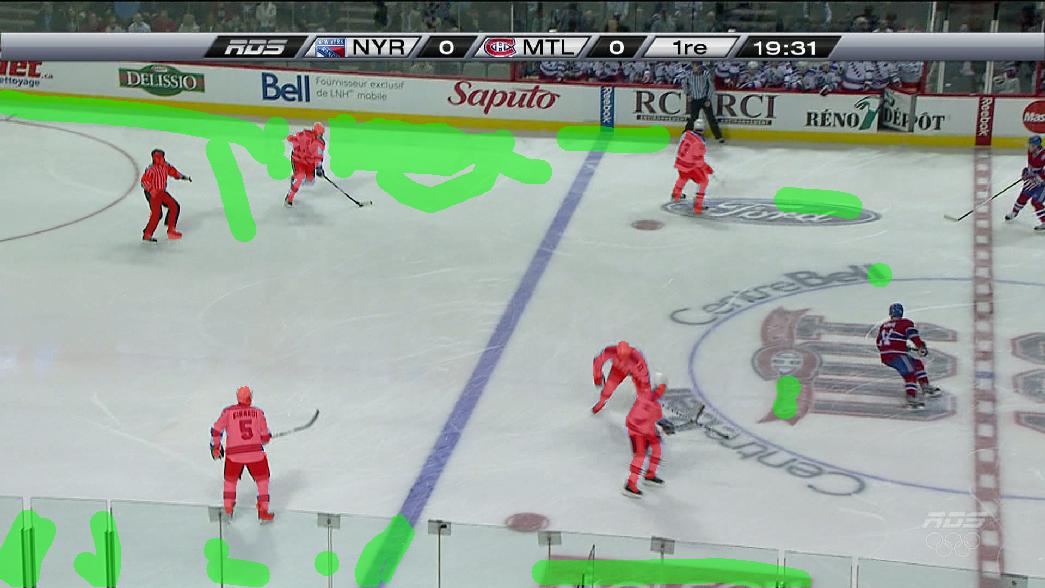}  &	
	\includegraphics[width=0.22\textwidth]{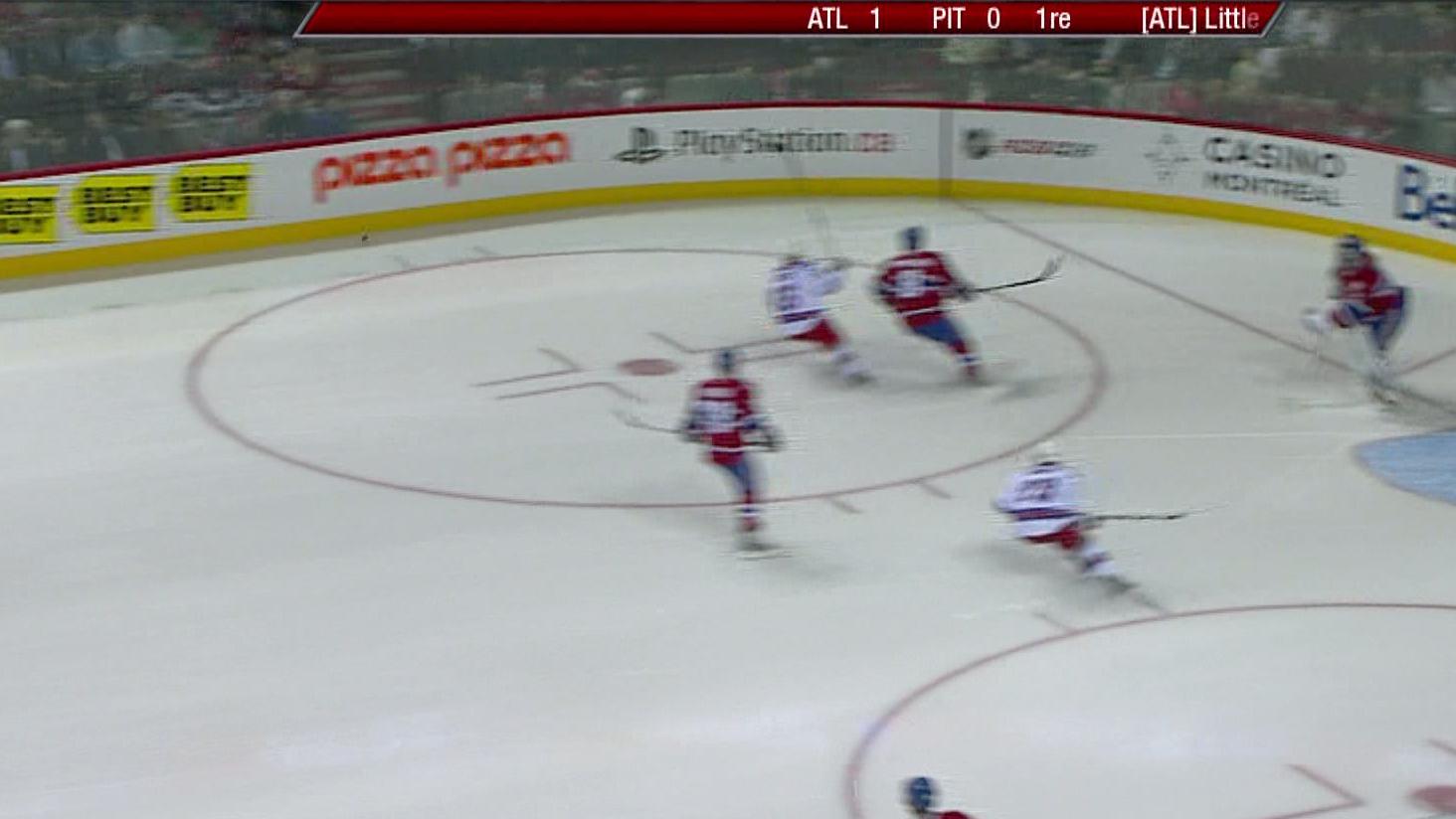}  &
	\includegraphics[width=0.22\textwidth]{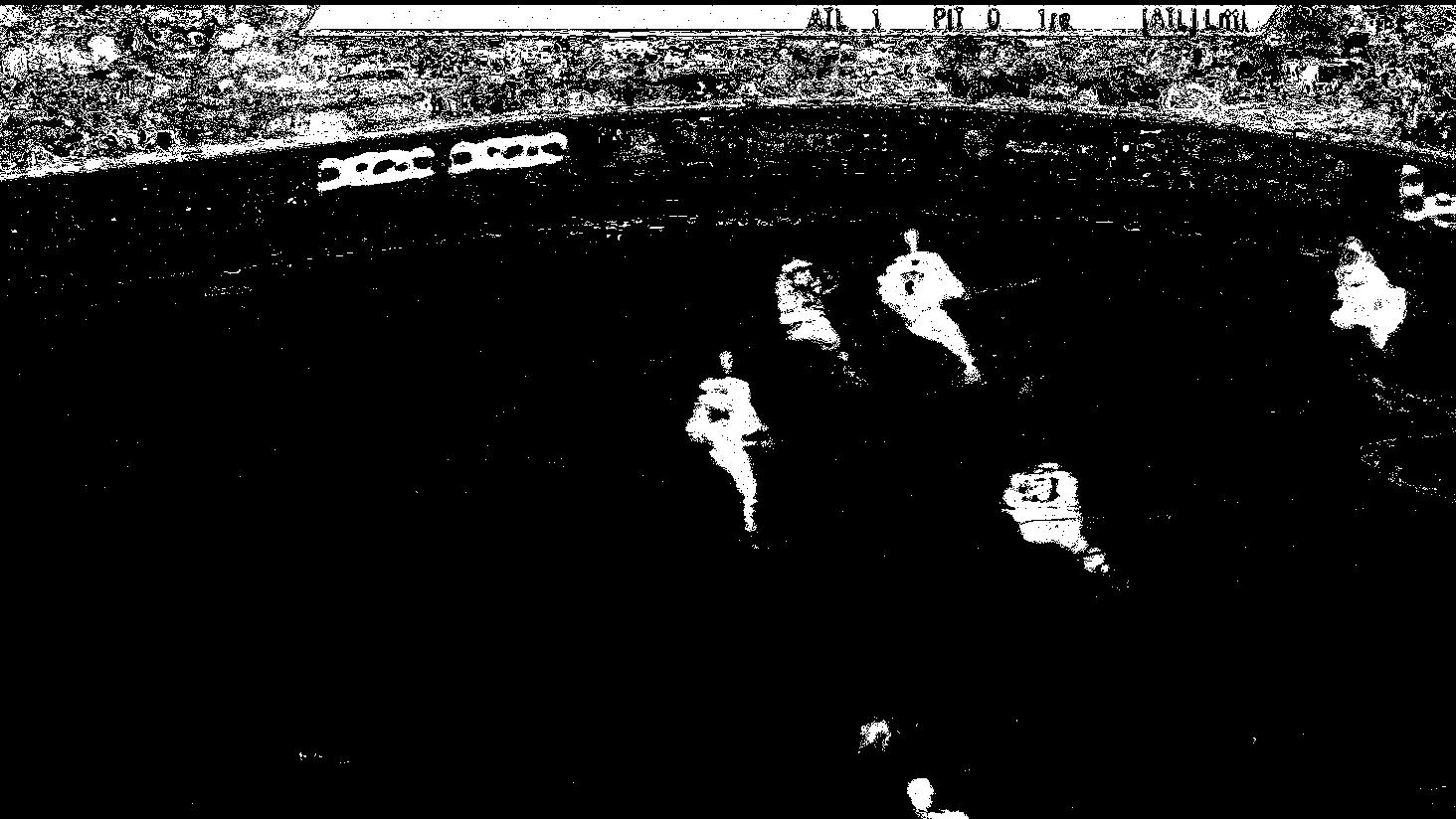}  \\
	\multicolumn{2}{c}{{\bf Training: interactively labelled pixels}} &
	\multicolumn{2}{c}{{\bf Test: frame 970}} 
\end{tabular}
  \caption[Figure-ground segmentation results]
	{{\bf Figure-ground segmentation results.}
	{\rm\quad  This shows results of figure-background segmentation on the hockey rink.  The left two images show the mask image of interactively labelled pixels where the red colour represents the foreground and the green colour represents the background.  The segmentation model is trained with 5 training images.  The right two images show the results of segmentation by the trained model.}}
\label{fig:segmentation_results}
\end{figure*}

Once detected players have their team label, the next step is to associate detected bounding windows into a set of ``tracklets'' where a tracklet represents a sequence of bounding windows that share the same identity over time. To achieve this, we employ a tracking-by-detection approach and adopt the tracking system of \citep{Lu2011} based on a Kalman filter \citep{Kalman_ASME1960}. In our self-learning process, we do not update parameters of a tracking model and treat player tracking as a black box.  Therefore, our system also works with other tracking-by-detection approaches such as a data-driven MCMC \citep{Khan2006} or the boosted particle filter (BPF) \citep{Okuma2004}.
\section{Data selection}
\label{sec:data_selection}

As described, a set of tracklets $\{\mathcal{T}\}_{j=1}^k$ is obtained by combining detection and tracking results of hockey players.  These tracklets are used as a pool of candidate data $\mathcal{C}$ from which we collect a set of training labels for improving performance of classification models.  Since this selection process is fully automatic, we need a selection criterion which effectively discovers additional training labels without accumulating incorrect labels.

Our selection criterion combines several image cues including detection, colour classification, tracking of players, and pixel-wise figure-ground segmentations.  The selection process is performed with the following steps.  First, we prune away short tracklets with less than 10 bounding windows because these tracklets are often produced by very sparse detection results, and often include incorrect labels.  After pruning, we have a refined set of tracklets $\{\mathcal{T}\}_{j=1}^m$ where $m < k$.  We initialize a pool of candidate data $\mathcal{C}$ with bounding windows of these tracklets.  Second, we compute the shape confidence of these predicted bounding windows by running our shape-based DPM detector on each bounding box. Third, we compute a foreground score $a_f \in [0, 1.0]$ to measure a proportion of foreground pixels (i.e., player pixels) within each predicted bounding window $\vect{R}_p$ in the candidate data $\mathcal{C}$:
\begin{equation}
	\label{eq:foreground_score}
    a_f = \frac{1}{area(\vect{R}_p)}\sum\limits_{\vect{d}_i\in\vect{R}_p}f(\vect{d}_i)
\end{equation}
where $area(\vect{R}_p)$ denotes the area of the bounding window $\vect{R}_p$ in terms of the total number of pixels within the window, and $f$ is a binary function which uses the decision value of our figure-ground segmentation model $H$ as follows: $f(\vect{d}_i) = 1$ if $H(\vect{d}_i) \geq 0$, or 0 otherwise.  We use a foreground score $a_f$ to determine whether or not the corresponding predicted bounding window $\vect{R}_p$ is added to a set of additional data $\mathcal{L}_{new}$.  For making this decision, we use labelled data and derive a set of two thresholds $\tau_{lower} = \mu_{a_f} - \sigma_{a_f}$ and $\tau_{upper} = \mu_{a_f} + \sigma_{a_f}$ where $\mu_{a_f}$ is a mean foreground score and $\sigma_{a_f}$ is a standard deviation.  These thresholds represent how likely $\vect{R}_p$ contains the foreground object in terms of the proportion of foreground pixels within the window and are computed based on all positive instances in ground-truth data.  Consequently, we add a predicted bounding window $\vect{R}_p$ to $\mathcal{L}_{new}$ if $\tau_{lower} \leq a_f \leq \tau_{upper}$.

The selected candidate data $\mathcal{L}_{new}$ is added to labelled data $\mathcal{L}$ by simply taking the union of these two datasets, $\mathcal{L} = \mathcal{L} \cup \mathcal{L}_{new}$.  This union produces many bounding windows that significantly overlap with each other.  We reduce these duplicates by prioritizing those instances in $\mathcal{L}_{new}$ and discarding existing instances in $\mathcal{L}$.  Assuming that classification models improve every iteration, we utilize this process for eliminating some of the incorrect localization labels.  However, such an assumption may not hold if the selection process accumulates too many noisy labels.  In the following experiments, we show that our assumption still holds in our self-learning framework.  
\begin{algorithm} [ht!]	
	\caption{{\bf: Data selection} \\ 
	 Given a set of tracklets $\{\mathcal{T}\}_{j=1}^k$ and a figure-ground segmentation model $H$, the goal is to select a portion of data as candidate labels for the next iteration of self-learning as described in \autoref{alg:self_learning}.  Every iteration, we set the maximum number of additional labels to be added as $n_{max}=2000$.}	
	\begin{algorithmic}[1]
		\STATE {{\bf Tracklet selection}: \\Discard short tracklets and initialize candidate data $\mathcal{C}$ from $\{\mathcal{T}\}_{j=1}^m$}.
		\STATE {{\bf Estimate the shape confidence of selected tracklets}: \\ 
		Run our shape-based DPM detector for each bounding window in $\mathcal{C}$. \\
		Sort them in ascending order of the predicted shape confidence.} 
		\STATE {{\bf Apply figure-ground segmentation}: \\ 
		For each bounding window $\vect{R}_p$, compute a segmentation score $a_f$ using \autoref{eq:foreground_score}.}
		\STATE {{\bf Final selection}: \\ 
		Select a new dataset $\mathcal{L}_{new}$ ($n_{max}$ additional labels) and merge datasets, $\mathcal{L}$: $\mathcal{L} = \mathcal{L} \cup \mathcal{L}_{new}$} 
	\end{algorithmic}			
	\label{alg:data_selection}
\end{algorithm}
\section{Experiments}
\label{sec:hockey_experiment}
\subsubsection*{Data}
Our system was tested on our hockey dataset consisting of 7 different video sequences which sum to 4,627 image frames of broadcast footage, and our basketball dataset consisting of 7 different video sequences which sum to 4,818 image frames of broadcast footage. The data are split into two separate sets: 3 sequences (2,249 frames in hockey, 2,486 frames in basketball) for training and 4 sequences (2,378 frames in hockey, 2,332 frames in basketball) for testing.  In the training data, the annotations are given in rectangular boxes with the category label, identification (i.e., the number of their jersey) and team colour label.  

In our experiments, we prepared 6 different sets of fully labelled images: 5 sets of $m$ randomly selected fully labelled images where $m = \{5, 10, 20, 40, 100\}$ and the fully supervised set of all 2,249 images for hockey and 2,486 images for basketball.  For each initial labelled dataset, we first trained the initial shape-based DPM detector and part-based colour classifiers.  Then we applied our self-learning framework to collect additional training labels from the unlabelled data and improve initial classifiers iteratively for up to four iterations. 
\subsubsection*{Player detection}
\begin{figure*}[ht!]
	\centering
	\begin{tabular}{cc}
	\includegraphics[width=0.45\textwidth]{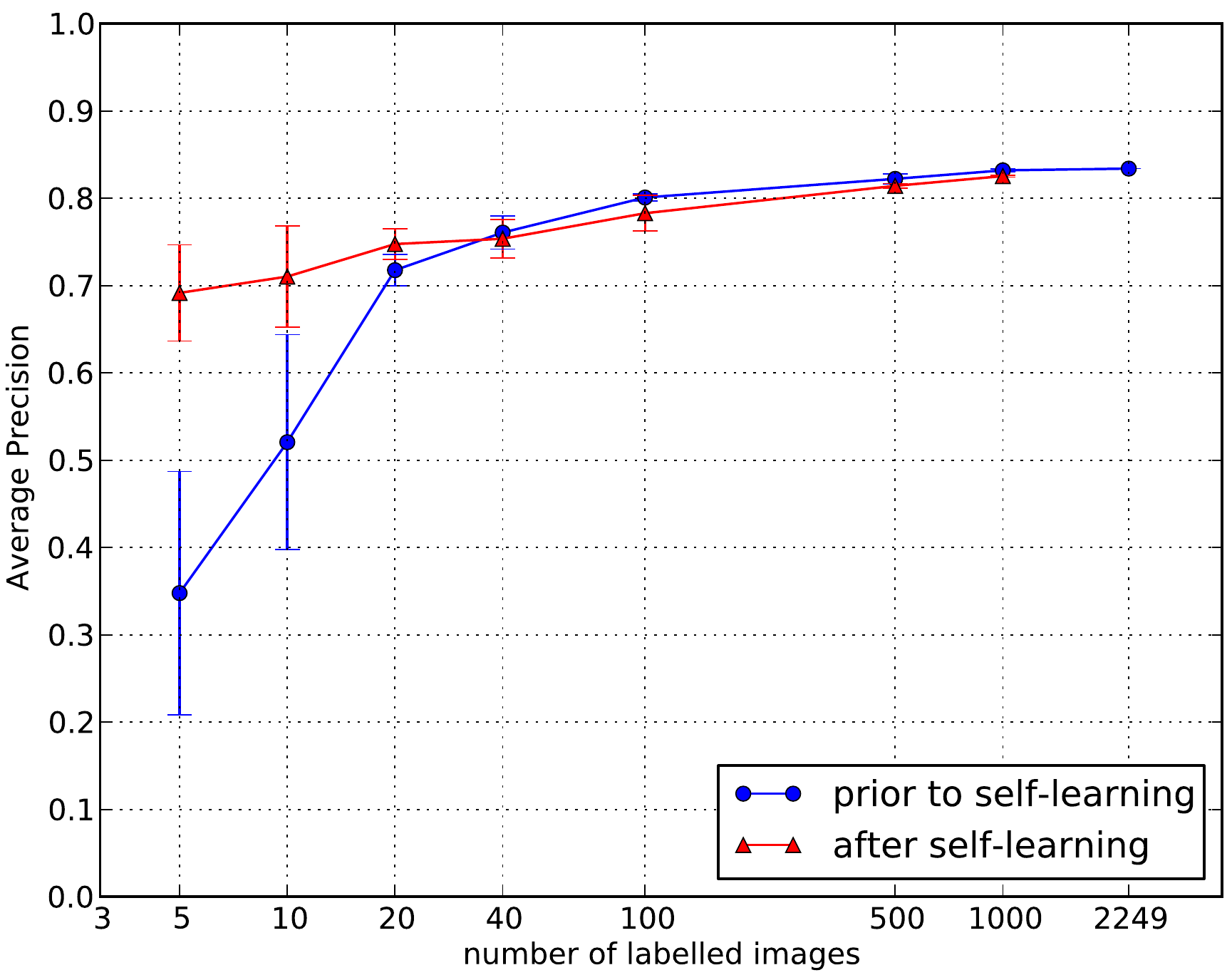} &
	\includegraphics[width=0.45\textwidth]{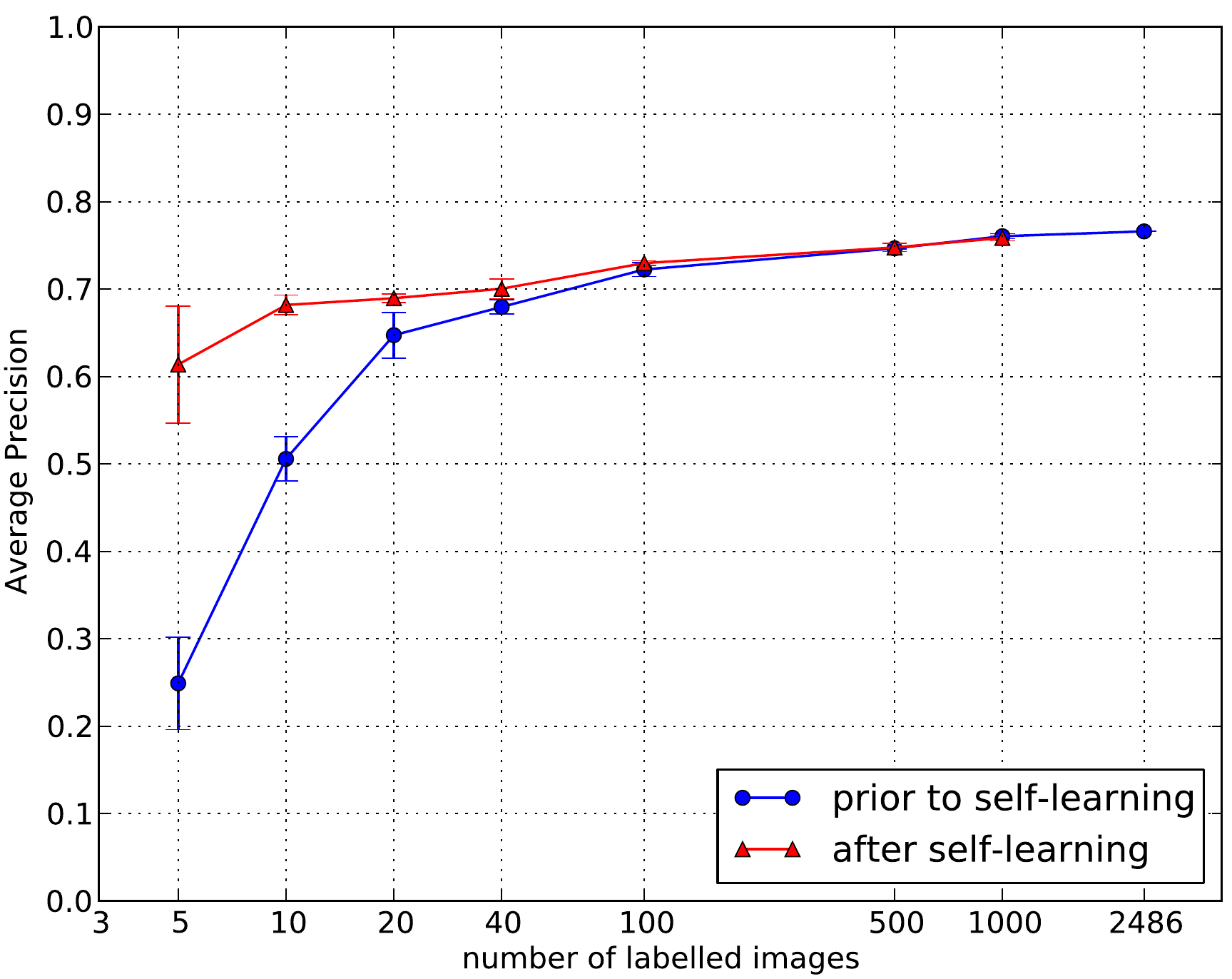} \\
	hockey & basketball 
	\end{tabular}
	\caption[Detection result of our weakly-supervised self-learning system in hockey and basketball videos]
	{{\bf Detection result of our weakly-supervised self-learning system in hockey and basketball videos.}
	{\rm\quad  The blue line shows the baseline performance based on only labelled datasets.  The red line shows the performance after four self-learning iterations of collecting additional labels from unlabelled data.  The x-axis is in a logarithmic scale.  Note a large performance gain when starting with as few as 5 labelled images.}}	
	\label{fig:small_result}
\end{figure*}

We adopted the PASCAL VOC criterion \citep{Everingham_IJCV2010} and used average precision (AP) for evaluating our detection results because it has been well defined and widely used in the vision community. 
\autoref{fig:small_result} shows the result of our system on our hockey data.  We ran the entire process five times and show the mean and variance for each labelled dataset.  The blue line shows the baseline performance based on only fully supervised data.  The red line shows the performance after our system collected additional labels from unlabelled parts of the video.  The results show a large performance gain --- about 20\% in the mean average precision --- in cases with a small number of labelled images (e.g., using 5 and 10 labelled images).  However, the performance gain gradually decreases or is eliminated with larger labelled datasets.   

\autoref{fig:small_result} shows the average number of labels used for each labelled dataset in the x-axis using a logarithmic scale.  We plot the average number of labelled bounding windows from each set of $m$ labelled images where $m = \{5, 10, 20, 40, 100\}$.  Note that each image typically contains multiple labels.

\subsubsection*{Player tracking}

\begin{figure*}[ht!]
	\centering
	\begin{tabular}{ c @{\hspace{1mm}} c }
		\includegraphics[width=0.45\textwidth]{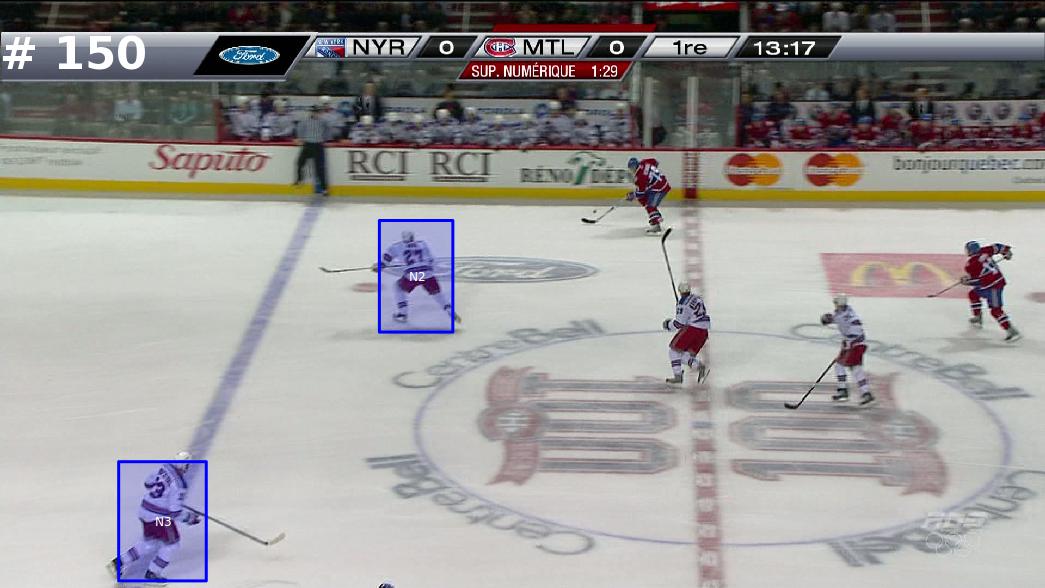}  &
		\includegraphics[width=0.45\textwidth]{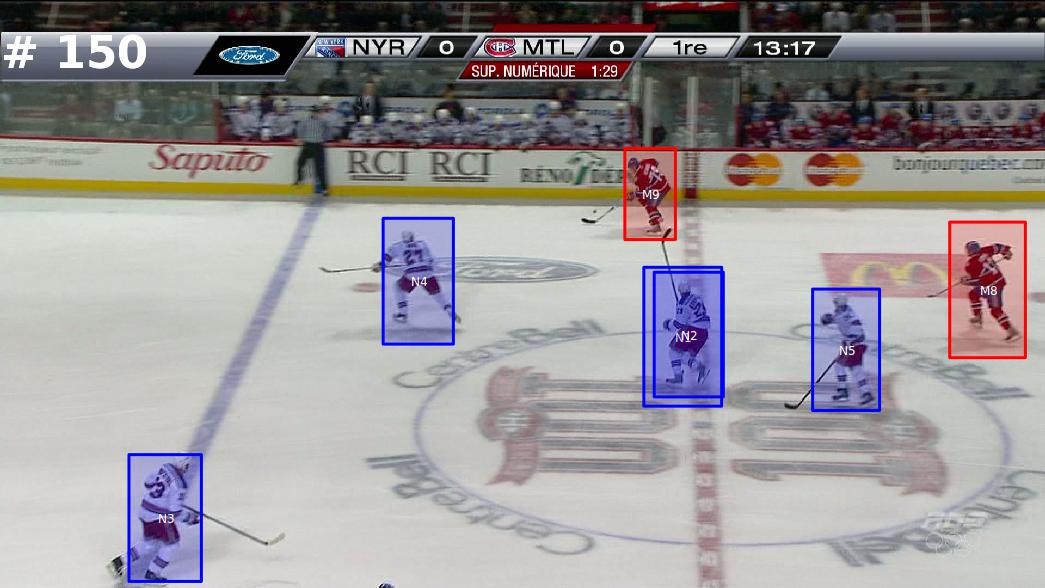}  \\
		\includegraphics[width=0.45\textwidth]{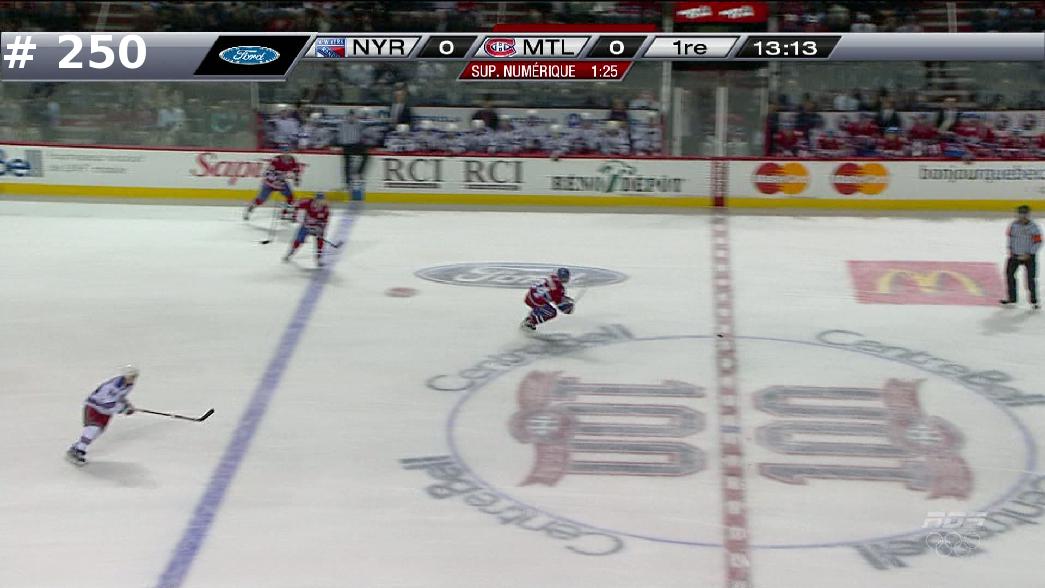}  &
		\includegraphics[width=0.45\textwidth]{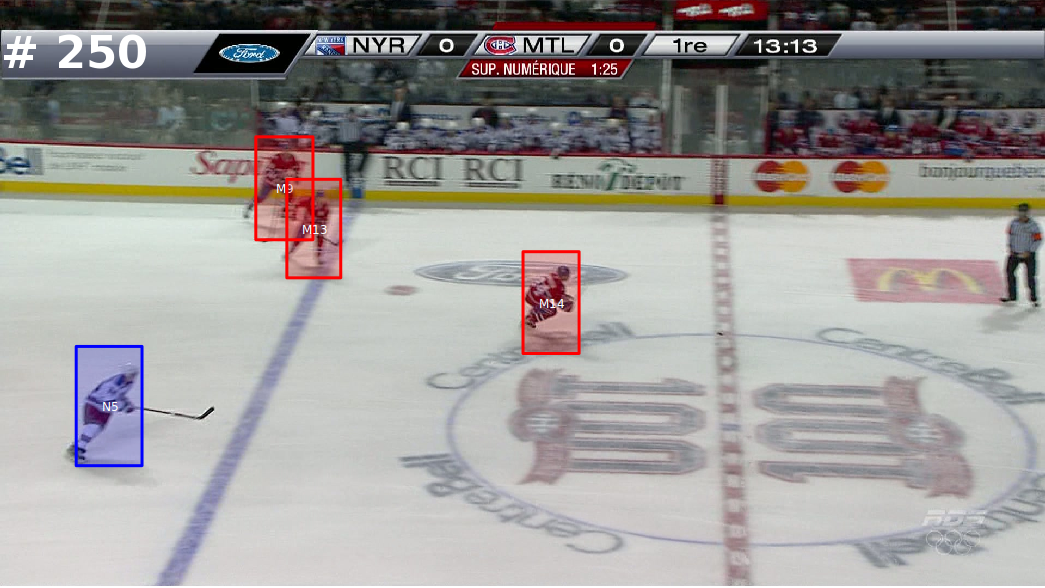}  \\
		\includegraphics[width=0.45\textwidth]{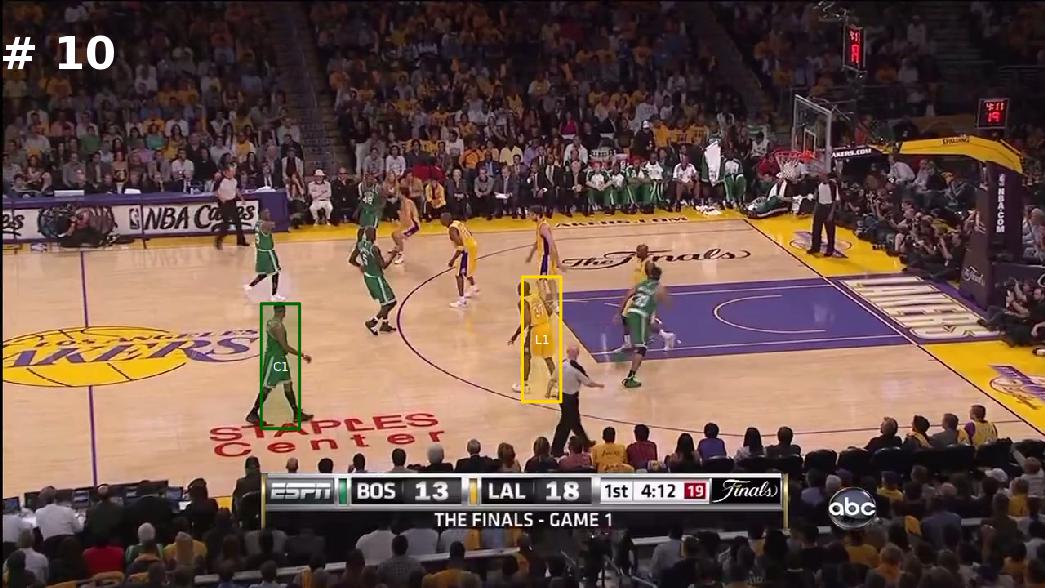}  &
		\includegraphics[width=0.45\textwidth]{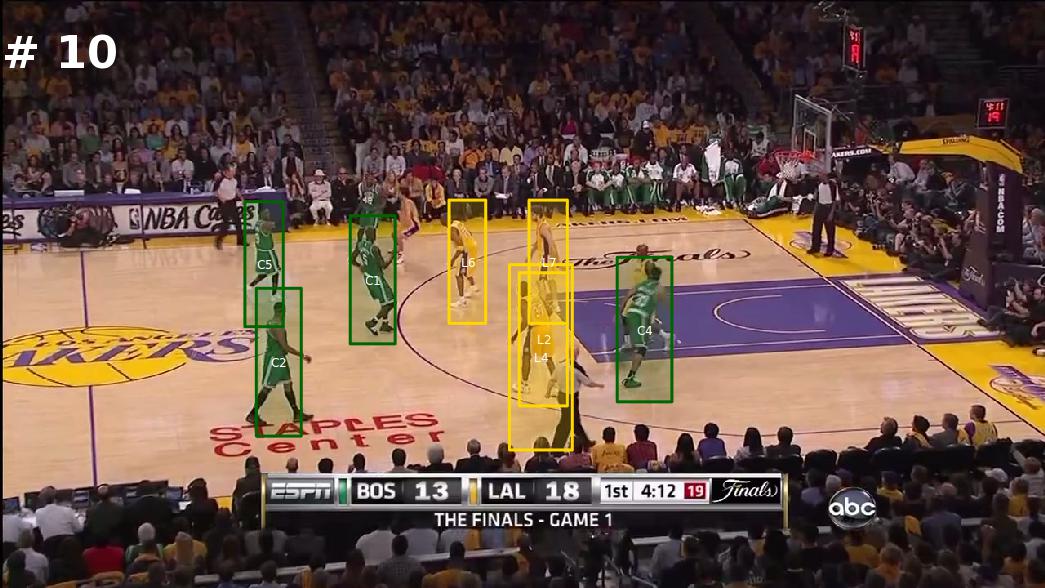}  \\
		\includegraphics[width=0.45\textwidth]{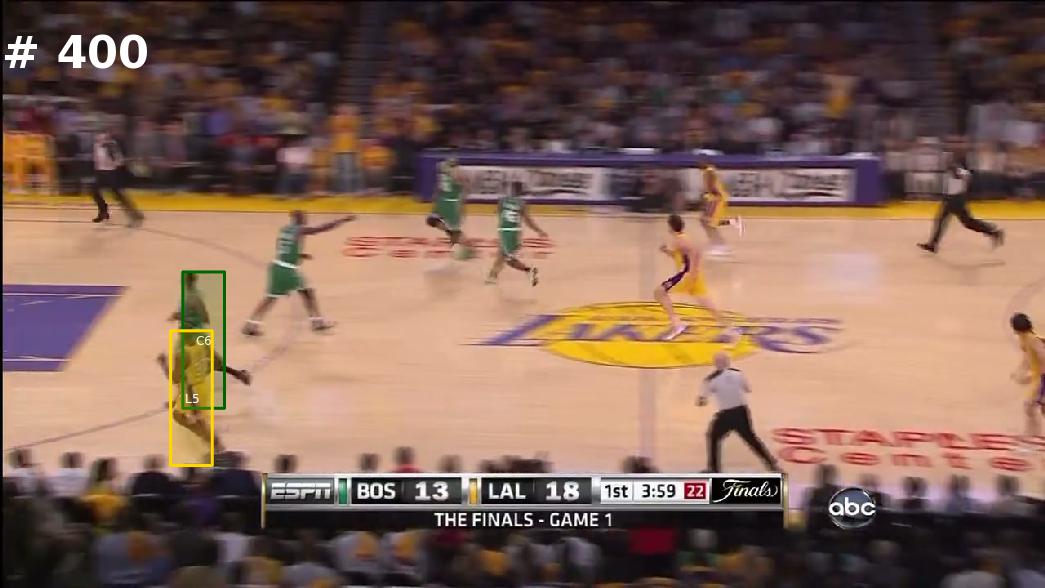}  &
		\includegraphics[width=0.45\textwidth]{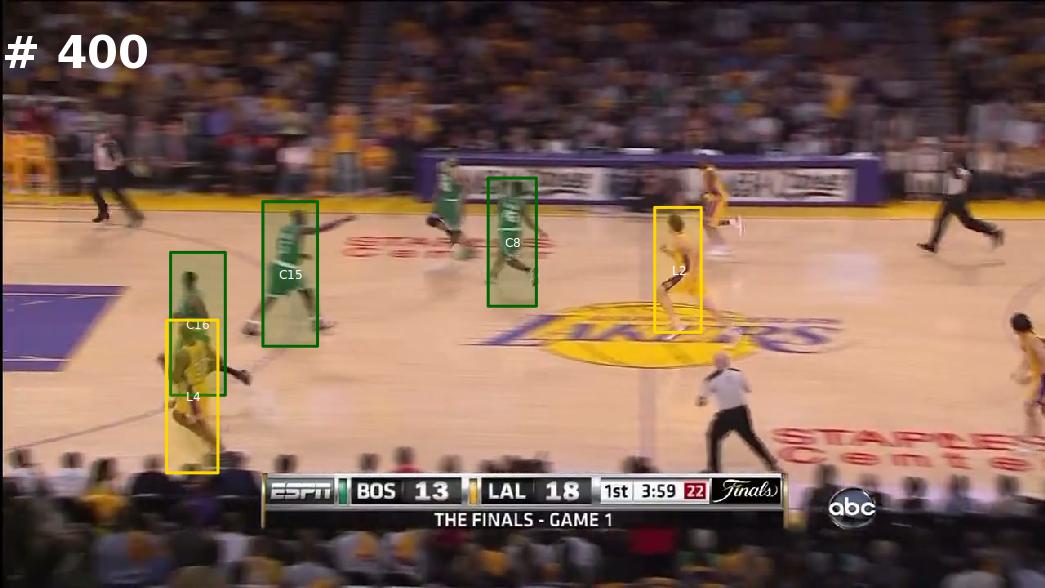}  \\		
		(a) Prior to self-learning & (b) After self-learning 
	\end{tabular}	
	\caption[Screenshots of our tracking result in sports videos]
	{{\bf Screenshots of our tracking result in sports videos.}
	{\rm\quad  This shows our tracking results on the test data. Column (a) uses detection inputs of a detector that is trained with 5 labelled images, which is the case of strongly supervised learning (SSL). Column (b) uses a detector that is trained with 5 labelled images as well as unlabelled data,  which is the case of weakly supervised learning (WSL).  Note that more players are discovered and tracked successfully after four self-learning iterations.  Short videos of these tracking results are available on YouTube: (1) For hockey: \url{http://youtu.be/uS0snd5fKi8}(2) For basketball: \url{http://youtu.be/_vyV2jbN5oc}}}	
	\label{fig:hockey_tracking_shots}
\end{figure*}

\autoref{fig:hockey_tracking_shots} shows the result of the weakly supervised training for 5 labelled images.  In the figure, more hockey players are discovered and tracked successfully after four self-learning iterations of our system in the case of 5 labelled images.  Secondly, the performance of tracking hockey players quickly converges to the best performance in the case of fully labelled images (e.g., compare one in 100 labelled images and one in fully labelled images). This fast convergence is also evident in the detection result of \autoref{fig:small_result}.

\subsubsection*{Data selection} 
\begin{figure*}[ht!]
	\centering
	\begin{tabular}{c | c | c}
		& most confident & least confident \\
		\hline
		{Iter 1 (.55)} & 
		\includegraphics[width=0.034\textwidth]{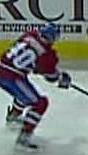} 
		\includegraphics[width=0.034\textwidth]{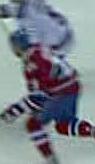}
		\includegraphics[width=0.034\textwidth]{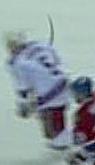}
		\includegraphics[width=0.034\textwidth]{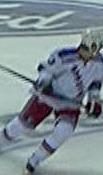}
		\includegraphics[width=0.034\textwidth]{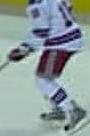} 
		\includegraphics[width=0.034\textwidth]{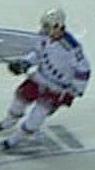} 
		\includegraphics[width=0.034\textwidth]{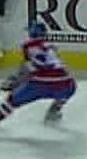}
		\includegraphics[width=0.034\textwidth]{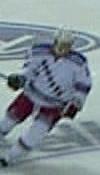}
		\includegraphics[width=0.034\textwidth]{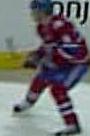}
		\includegraphics[width=0.034\textwidth]{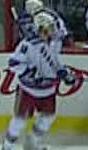} &
		\includegraphics[width=0.034\textwidth]{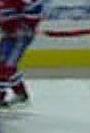} 
		\includegraphics[width=0.034\textwidth]{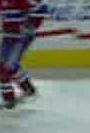}
		\includegraphics[width=0.034\textwidth]{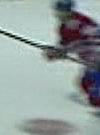}
		\includegraphics[width=0.034\textwidth]{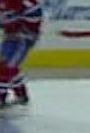}
		\includegraphics[width=0.034\textwidth]{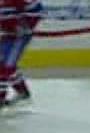} 
		\includegraphics[width=0.034\textwidth]{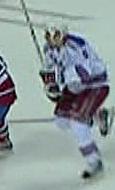} 
		\includegraphics[width=0.034\textwidth]{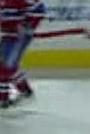}
		\includegraphics[width=0.034\textwidth]{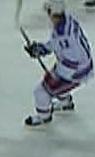}
		\includegraphics[width=0.034\textwidth]{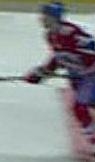}
		\includegraphics[width=0.034\textwidth]{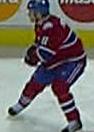} \\
		\hline
		{Iter 2 (.62)} &
		\includegraphics[width=0.034\textwidth]{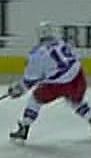} 
		\includegraphics[width=0.034\textwidth]{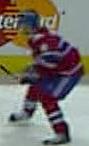}
		\includegraphics[width=0.034\textwidth]{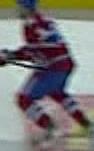}
		\includegraphics[width=0.034\textwidth]{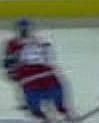}
		\includegraphics[width=0.034\textwidth]{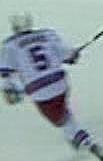} 
		\includegraphics[width=0.034\textwidth]{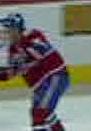} 
		\includegraphics[width=0.034\textwidth]{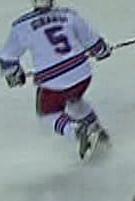}
		\includegraphics[width=0.034\textwidth]{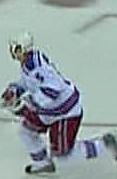}
		\includegraphics[width=0.034\textwidth]{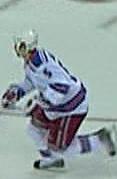}
		\includegraphics[width=0.034\textwidth]{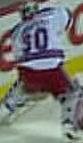} &		
		\includegraphics[width=0.034\textwidth]{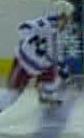} 
		\includegraphics[width=0.034\textwidth]{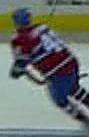}
		\includegraphics[width=0.034\textwidth]{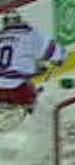}
		\includegraphics[width=0.034\textwidth]{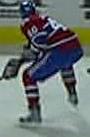}
		\includegraphics[width=0.034\textwidth]{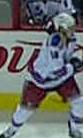} 
		\includegraphics[width=0.034\textwidth]{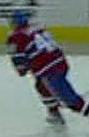} 
		\includegraphics[width=0.034\textwidth]{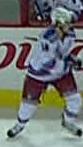}
		\includegraphics[width=0.034\textwidth]{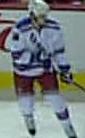}
		\includegraphics[width=0.034\textwidth]{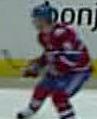}
		\includegraphics[width=0.034\textwidth]{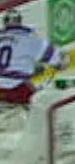} \\
		\hline
		{Iter 3 (.64)} &
		\includegraphics[width=0.034\textwidth]{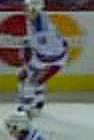} 
		\includegraphics[width=0.034\textwidth]{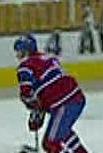}
		\includegraphics[width=0.034\textwidth]{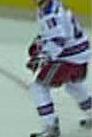}
		\includegraphics[width=0.034\textwidth]{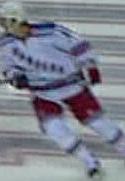}
		\includegraphics[width=0.034\textwidth]{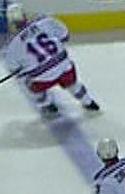} 
		\includegraphics[width=0.034\textwidth]{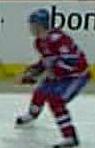} 
		\includegraphics[width=0.034\textwidth]{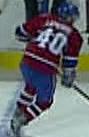}
		\includegraphics[width=0.034\textwidth]{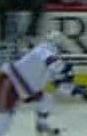}
		\includegraphics[width=0.034\textwidth]{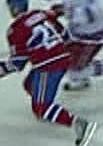}
		\includegraphics[width=0.034\textwidth]{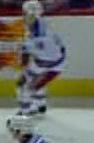} &	
		\includegraphics[width=0.034\textwidth]{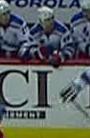} 
		\includegraphics[width=0.034\textwidth]{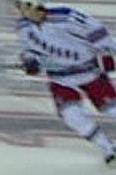}
		\includegraphics[width=0.034\textwidth]{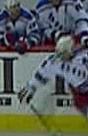}
		\includegraphics[width=0.034\textwidth]{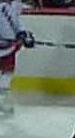}
		\includegraphics[width=0.034\textwidth]{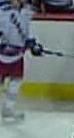} 
		\includegraphics[width=0.034\textwidth]{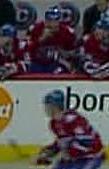} 
		\includegraphics[width=0.034\textwidth]{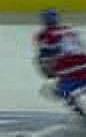}
		\includegraphics[width=0.034\textwidth]{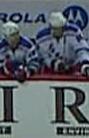}
		\includegraphics[width=0.034\textwidth]{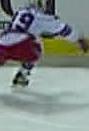}
		\includegraphics[width=0.034\textwidth]{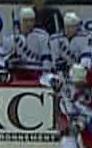} \\		
		\hline
		{Iter 4 (.67)} &
		\includegraphics[width=0.034\textwidth]{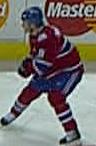} 
		\includegraphics[width=0.034\textwidth]{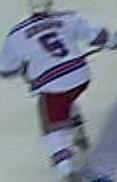}
		\includegraphics[width=0.034\textwidth]{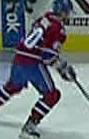}
		\includegraphics[width=0.034\textwidth]{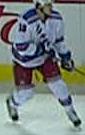}
		\includegraphics[width=0.034\textwidth]{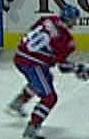} 
		\includegraphics[width=0.034\textwidth]{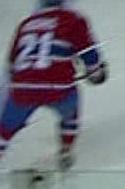} 
		\includegraphics[width=0.034\textwidth]{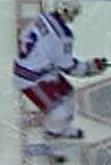}
		\includegraphics[width=0.034\textwidth]{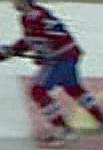}
		\includegraphics[width=0.034\textwidth]{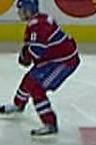}
		\includegraphics[width=0.034\textwidth]{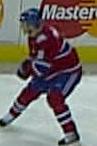} &		
		\includegraphics[width=0.034\textwidth]{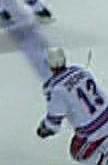} 
		\includegraphics[width=0.034\textwidth]{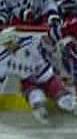}
		\includegraphics[width=0.034\textwidth]{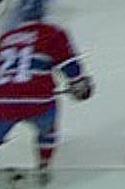}
		\includegraphics[width=0.034\textwidth]{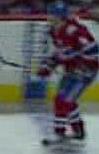}
		\includegraphics[width=0.034\textwidth]{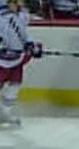} 
		\includegraphics[width=0.034\textwidth]{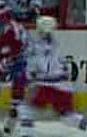} 
		\includegraphics[width=0.034\textwidth]{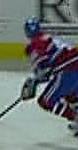}
		\includegraphics[width=0.034\textwidth]{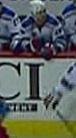}
		\includegraphics[width=0.034\textwidth]{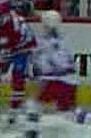}
		\includegraphics[width=0.034\textwidth]{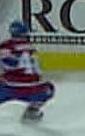} \\
		\hline			
	\end{tabular}
	\caption[Most confident and least confident candidate bounding windows in hockey videos]
	{{\bf Most confident and least confident candidate bounding windows in hockey videos.}
	{\rm\quad  This shows the most confident (i.e., highest scoring detection) and the least confident (i.e., lowest scoring detection) candidate bounding windows that are selected from unlabelled images in the training data by Algorithm \autoref{alg:data_selection}.  The average precision of our detection model on the test data for each iteration is shown in the parentheses.  Note how the localization of hockey players is improved from the iteration 1 to 4.}}
	\label{fig:hockey_candidate}
\end{figure*}

\autoref{fig:hockey_candidate} shows representative candidate bounding windows in each iteration of the self-learning process.  The figure shows the most confident bounding windows with a high detection score and the least confident bounding windows with a low detection score among candidate bounding windows that are selected by our data selection algorithm \autoref{alg:data_selection}.  The localization of hockey players is improved gradually in each iteration.  The difference is especially obvious between the iteration 1 and 4, where there is an improvement of 12\% in the average precision.  Importantly, many of these candidate bounding windows are typically false negatives of the player detector.  The detector alone cannot identify these misclassification examples, but they are quite effective at improving the classification performance \citep{Okuma2011}.  Our approach is able to select them by tracking players' motions and segmenting the colour of the playing field. 
\subsubsection*{Computation time:}
Our experiments were performed on an 8-core (Intel Xeon 2.66GHz) machine with 32GB of RAM.  The weakly supervised case had four additional learning iterations on top of the strongly supervised case which required only one iteration for training and testing.  It took about 4 days of CPU time to run our system on all labelled datasets, where over 80\% of time was spent for training a detector and running it on both training and test images to obtain detection bounding windows.  It takes about 7 to 10 seconds to run our DPM detector on an image of $960 \times 540$.  To speed up the detection process, the size prior of sports players was estimated from training data and used to focus computational resources within a limited range of scales --- in our case, $[\mu_s-\sigma_s, \mu_s+\sigma_s]$ where $\mu_s$ is the mean size and $\sigma_s$ is a standard deviation.  

\section{Conclusions}
Our self-learning approach combines several image cues such as the appearance information (i.e., shape and colour) of players, the constraints on their motions, and the colour of the playing field for discovering additional labels automatically from unlabelled data.  We use the constraints of players' motions to explore unlabelled portions of sports videos and discover useful labels that the appearance-based player detector is unable to find with the current classification performance.  The playing field segmentation is effective for eliminating erroneous labels.  Our experimental results show that our approach is particularly effective when there is very little labelled data.   

This paper shows that it is possible to realize fully automatic acquisition of labels if a small amount of label data is available even in realistic, challenging videos from broadcast footage of sports.  An immediate future direction is to use a game-specific player detector for re-targeting other games (e.g., classic games that have been recorded in the past) by re-learning the confidence score of the detector without additional manual labels as in \citep{Wang2012}.  Ideally, the label acquisition process should be fully automatic, which will be a difficult goal to achieve in general.  Although we showed the possibilities in sports video, there are still many challenges that need to be resolved in order to realize fully automatic acquisition of labels for solving the problem of generic object detection.

\bibliographystyle{spbasic}      
\bibliography{mylibrary}   

\end{document}